# "The Taurus": Cattle Breeds & Diseases Identification Mobile Application using Machine Learning

R.M.D.S.M.Chandrarathna[1], T.W.M.S.A.Weerasinghe[2], N.S.Madhuranga[3], T.M.L.S.Thennakoon[4], Anjalie Gamage[5] and Erandika Gamage[6]

[1]Faculty of Computing, Sri Lanka Institute of Information Technology, (SLIIT), Malabe, SRI LANKA
[2]Faculty of Computing, Sri Lanka Institute of Information Technology, (SLIIT), Malabe, SRI LANKA
[3]Faculty of Computing, Sri Lanka Institute of Information Technology, (SLIIT), Malabe, SRI LANKA
[4]Faculty of Computing, Sri Lanka Institute of Information Technology, (SLIIT), Malabe, SRI LANKA
[5]Faculty of Computing, Sri Lanka Institute of Information Technology, (SLIIT), Malabe, SRI LANKA
[6]Faculty of Computing, Sri Lanka Institute of Information Technology, (SLIIT), Malabe, SRI LANKA

[1]Corresponding Author: surekhamaduhansi@gmail.com

## ABSTRACT

Dairy farming plays an important role in agriculture for thousands of years not only in Sri Lanka but also in so many other countries. When it comes to dairy farming cattle is an indispensable animal. According to the literature surveys almost 3.9 million cattle and calves die in a year due to different types of diseases. The causes of diseases are mainly bacteria, parasites, fungi, chemical poisons and etc. Infectious diseases can be a greatest threat to livestock health. The mortality rate of cattle causes a huge impact on social, economic and environmental damage. In order to decrease this negative impact, the proposal implements a cross-platform mobile application to easily analyze and identify the diseases which cattle suffer from and give them a solution and also to identify the cattle breeds. The mobile application is designed to identify the breeds by analyzing the images of the cattle and identify diseases after analyzing the videos and the images of affected areas. Then make a model to identify the weight and the age of a particular cow and suggest the best dose of the medicine to the identified disease. This will be a huge advantage to farmers as well as to dairy industry. The name of the proposed mobile application is "The Taurus" and this paper address the selected machine learning and image processing models and the approaches taken to identify the diseases, breeds and suggest the prevention methods and medicine to the identified disease.

## I. INTRODUCTION

Cattle are considered as the most useful domestic animal in the world. We have so many evidences which says that, cattle are used for so many works throughout the ancient times and cattle have contributed to the survival of human for many thousand years [1]. They have contributed in human welfare supplying, milk powder, milk, gee and so many other dairy products which are enriched with high nutrition. Having those helps in making good heart health, boots up immune system, prevent humans from various types of diseases such as diabetes [2].

Apart from the benefits which humans take from the cattle, there are so many other benefits to the environment also. Cattle is the number one agricultural source of greenhouse gases worldwide. Proper cattle grazing management can help to mitigate climate changes and also cows help in restoring healthy soils, conserve sensitive species and enhancing overall ecological function [3]. But most of the time people fail to appreciate or recognize the advantages of having cattle on earth. Each and every country in the world get the advantages of cattle. If cows inhabit the earth, human would no longer have access to milk or countless dairy products which it spawns. Globally, dairy provides 5% of the energy in diet. So having no cattle means that humans would lose a key source of minerals and vitamins including calcium, phosphorus, zinc, potassium, vitamin A and D, etc. [4].

When it comes to the reasons of the decrement of cattle population, there are reasons such as respiratory problems, digestive problems, and diseases as the leading reasons [5]. In most countries like Sri Lanka, most of the farmers and cattle owners own a small-scale farm. And most of them do not have a proper way to manage the diseases which their cattle get diagnosed with. In some areas it is very much difficult to contact a veterinarian, so that the diseased cattle would have dead by the time the veterinarian arrives. And most of the time the owners don't have a way to take their animal to get the medical attention. Because of this reason the farm owners lose the main way of their income.

Considering these reasons this study explores the use of ICT solutions to implement a mobile application to easily identify and get to know about the cattle breeds and diseases and let the user know the prevention methods. Today in the modern world, each and every person owns a smart phone. Therefore, considering the difficulties cattle owners face when treating their sick cows, having a mobile application which can identify the breeds, diseases and prevention methods can lead the farmers to save most of their sick cows with a less help of veterinarians and within a less time. This will lead the increment of cattle population and the economy of the farmers.

## II. BACKGROUND AND LITERATURE SURVEY

When identifying the diseases of cattle, their breed also plays an important role. In the world there are 250+ cattle breeds. Cattle can be used for many purposes therefore cattle need to be saved and protected. There are mainly dairy cattle breeds and beef cattle breeds. In our proposed system we concern about dairy cattle breeds which is a main resource in the society. Thus, dairy cattle act as an important factor in dairy industry. Though farmer in the closest person who deal with cattle, they are aware about only few breeds which are the breeds they are dealing in day-to-day activities. And others who concern about cattle such as students, doctors, people who are interested about cattle etc.

There are mainly 2 cattle breeds they are dairy cattle breeds and beef cattle , though countries like Sri Lanka based on only dairy production our European countries are interested in both dairy and beef both. Accordingly, those two breeds are having different characterizes in between two breeds. The genetic diseases in dairy and beef cattle are tissue specific. There is difference in occurring generic diseases such as below figure.

**Table I:** Generic diseases

| Sr. No. | Specific tissue | Dairy Cattle | Beef Cattle |
|---|---|---|---|
| 1 | Skeletal | a. Chrondrodysplacia<br>b. Complex vertebral malformation<br>c. Osteogenesis Imperfecta<br>d. Osteoporosis<br>e. Syndactylism | a. Osteoporosis<br>b. Arachnomelia<br>c. Arthrogyposis Multiplex<br>d. Congenital Contractual Arachnodactyly<br>e. Syndactylism |
| 2 | Central Nervous System | a. Weaver Syndrome<br>b. Spinal Dysmielination<br>c. Spinal Muscular Atrophy | a. Idiopathic Epilepsy<br>b. Neuronal Ceroid Lipofuscinosis<br>c. Hydrocephalus<br>d. Spastic Paresis |
| 3 | Blood | a. BLAD<br>b. Hereditary Zinc Deficiency<br>c. Citrullinemia | Nil |
| 4 | Skin | a. Epitheliogenesis Imperfecta<br>b. X-Linked Anhidrotic Ectodermal | a. Hypotrichosis |
| 5 | Muscle | | a. Congenital |
| | Function Disorder | Nil | Pseudomyotonia<br>b. Crooked tail Syndrome |
| 6 | Ophthalmic | a. Anapthlmos and Microthlmos<br>b. Congenital Cataract<br>c. Optic Nerve Colobomas | a. Anapthlmos and Microthlmos<br>b. Congenital Cataract<br>c. Optic Nerve Colobomas |

The various scientists had reported the frequency as 3.33% and 4.0% in Iranian Holstein Friesian and Chinese Holstein cattle, respectively. However, in India BLAD carrier was estimated as 3.23% in pure and crossbred Holstein-Friesian only. The details of these genetic diseases with special reference to its definition, genetic cause (DNA mutation) and its clinical symptoms are discussed in the review [6]. Therefore, identifying the breed of a cattle is one of the most crucial factors where there to identify any generic diseases for that specific cattle breed and provide necessary treatments.

In Sri Lanka, there are 03 categories as, up-country, low –country and mid country, as such they are as follows:

**Table II:** Cattle breeds in Sri Lanka

| Up-country | • Ayrshire<br>• Jersey<br>• Friesian |
|---|---|
| Low-country | • Sindhi<br>• Sahiwal<br>• Tharparkar<br>• AMZ(Australian Milking ZEBU)<br>• AFS(Australian Friesian Sahiwal)<br>• Local Crossbreeds |
| Mid-country | • Jersey<br>• Friesian<br>• AMZ |

Though there are comparably a smaller number of breeds in Sri Lanka there are cross breeds also, and other main breeds are also can be seen all around Sri Lanka. Thus, they have many more special characteristics with their living environment, climate etc. [7]. To understand their special characteristics and their features identifying the breed is an important task which leads to provide a best treatment.

According to the above reading I have done, understanding a breed is a main task and it makes the treatment a best, which the cattle can be treated by identifying the breed and they're by understanding their breed generic disease. As such by understanding and implementing a procedure to identify a cattle breed it will be an great opportunity to protect our cattle population.

When it comes to the diseases, there are various diseases which are diagnosed by cattle. Following is some of them.

**Table III:** Cattle diseases

| No | Cattle disease | Number of countries that have a CP in place |
|---|---|---|
| 1 | Enzootic Bovine Leucosis (EBL) | 31 |
| 2 | Bluetongue | 27 |
| 3 | Infectious Bovine Rhinotracheitis (IBR) | 24 |
| 4 | Bovine Viral Diarrhea (BVD) | 23 |
| 5 | Anthrax | 16 |
| 6 | Par tuberculosis | 15 |
| 7 | Salmonellosis | 8 |
| 8 | Salmonellosis | 7 |
| 9 | Leptospirosis | 7 |
| 10 | Trichomonosis | 7 |
| 11 | Neosporosis | 6 |
| 12 | Liver fluke | 5 |
| 13 | Streptococcal infection | 5 |
| 14 | Q fever | 4 |
| 15 | Aujeszky's disease | 4 |
| 16 | Mycoplasmosis | 3 |
| 17 | Contagious bovine pleuropneumonia | 2 |
| 18 | Staphylococcal infection | 2 |
| 19 | Bovine respiratory disease | 2 |
| 20 | Epizootic haemorrhagedisease | 1 |
| 21 | coronavirus inflation | 1 |
| 22 | Ringworm | 1 |
| 23 | Bovine digital dermatitis | 1 |

From the above diseases foot and mouth disease, mastitis, lumpy skin disease can be identified using image data.

For an example Mastitis disease is caused by Streptococcus bacteria and milk production is reduced due to this disease. Also, if not treated at the right time, the udder will be damaged and milk production will not be restored. The symptoms of this are udder such as swelling, redness, heat, hardness, pain and the milk such as watery appearance and flakes. Often soon after calving, with the abnormal milk having to be discarded and inject some antibiotic therapy, disease can be prevented.

One of the main factors to check whether a person is sick, is the particular person's behavior change. Not only for humans is this applicable for animals also. According to the research which has been done so far, the identification of sick dairy cows in an early state of the disease can be done by observing the individual cow's behavior. Other than that, observing behavioral changes plays a significant role in identifying the injuries and diseases the cattle suffer from.

Cattle thereby exhibit distinctive behavioral patterns such as walking, resting, lying, feeding, social behavior and other physical activities which are usually changed by other external factors such as climate changes and diseases. Bovine Spongiform Encephalopathy, Lameness, Heat Stress can be taken as main diseases which result behavioral changes [8].

The following figure shows the main changes in behavior when a cow is diagnosed with mad cow disease or the Bovine spongiform encephalopathy.

**Table IV:** Changes in behavior of cattle with mad cow disease

| Changes in behavior patterns | Changes in locomotion | Sensory disturbances |
|---|---|---|
| • General apprehension<br>• Tremors-Head/Neck /whole body<br>• Head shy<br>• Aggressive<br>• Nervous of doorways<br>• Increased vocalization<br>• Social behavior<br>• Fine muscle fascination<br>• Hyperresponsive to touch/sound | • Hindlimb ataxia<br>• Falling<br>• Abnormal posture<br>• Forelimb ataxia<br>• Recumbency<br>• Circling | • Head tossing<br>• Abnormal ear movement<br>• Teeth grinding<br>• Nose licking<br>• Tongue play<br>• Head rubbing/pressing<br>• Sneezing/snorting<br>• Yawning |

After identifying the diseases, the main thing to do to save the cattle population is to suggest the treatments. The given medicine dosage differs from the age and the weight of the cattle.

Before provide medicine for animals we need to identify best dose by considering age, weight, gender, etc. Here we can identify age of the cattle by capture and upload images to the system of diseased cattle's mouth to estimate cattle age using image processing technique [9].

The medicine dosage is always based on the "Age" and "Weight". So that we need to detect the "Age" and "Weight" of diseased cattle before suggesting the dosage of medicine [10].

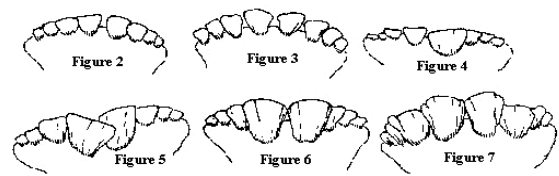

**Figure 1:** Dentition according to the age

**Table V:** Age of cow based on dentition

| Actual Age | Teeth Present | Other Comments |
|---|---|---|
| Less than 2 years old | Two permanent incisors present | - |
| 2 years old | Four permanent incisors present | Will be the middle incisor teeth- called pincers |
| 3 years old | Six permanent incisors present | Called first intermediate- One |

| | | on each side of pincers |
|---|---|---|
| 4 years old | Eight permanent incisors present | Called second intermediate |
| 5 years old | Ten permanent incisors present | Called corner incisor |
| Older than 6 years | All teeth present | Age based on tooth wear separation and visibility of tooth root |
| About 12 years | Some may be missing | The arch in mouth has disappeared and teeth become triangular with extremely noticeable wear |

Another important fact which we consider before treatment is cattle's' "Weight".

Because when we suggest medicine for diseased cattle, we should always know the weight of the cattle. Also need to constantly weigh animals to check whether **they are healthy or** not. The data collected after weighing them will give you a better understanding of when to increase or reduce their feed. The data can also be used to suggest any changes in the diet or recommend more nutrients.

Through image processing the weight and the age of cattle can be identified.

## III. METHODOLOGY

When considering the overall process of "The Taurus" mobile application, it consists of four main functions, named as breeds identification, disease identification using images, disease identification using videos and predict the best medicine dosage for an identified disease considering age and the weight of the cow.

Identifying a cow's breed is a critical factor in identifying the genetic diseases. In the present world there are so many small-scale farms which has no solution to identify the cattle breeds, in the process of identifying the diseases they may face. Therefore, this study aims to give an ICT solution to identify cattle breeds in the process of identifying the diseases faced by cattle. Accordingly, we found that special features between the cattle breeds and they can be differentiated using the special characteristics and identify them using them. Features such as tail, body shape, horns differentiate cattle breeds from each other.

In the process of training the dataset the MobileNet V2, which is an efficient convolutional neural network for mobile vision is used.

**Table VI:** Details regarding breeds image dataset

| Identified Breed | Number of images trained |
|---|---|
| Ayrshire cattle breed | 260 |
| Brown Swiss cattle breed | 238 |
| Holstein Friesian cattle breed | 254 |
| Jersey cattle breed | 252 |
| Unknown | 119 |

The dataset for the component is collected through websites, small scale farms in Sri Lanka and Abewela Farm Sri Lanka.

Identifying the disease by uploading an image is another solution we have implemented in the study. According to the surveys we have done, the unavailability of a mobile function to identify diseases using machine learning techniques is identified and here in our study we gave a solution for the unavailability.

In the process of identifying the diseases also the CNN's (Convolutional Neural Network) approach, Mobile Net V2 is used. DiseasesIdentification.h5 model was created using Mobile Net V2 which is a very effective algorithm to feature extraction for object detection and segmentation. Following is the details of Identified diseases and number of data trained.

**Table VII:** Details regarding diseases image dataset

| Identified Disease | Number of images trained |
|---|---|
| Bovine John's Disease | 32 |
| Foot and mouth disease | 75 |
| Lumpy skin disease | 92 |
| Mastitis disease | 74 |
| Milk fever disease | 28 |
| Healthy cow | 61 |
| Unknown | 92 |

Identifying cattle's behavior is also a critical factor in identifying the diseases cattle may face. There are no mobile applications available to identify the diseases of a cow considering the behavioral changes. Therefore, one of the aims of this study is to give an ICT solution to analyze the changes of the normal behavior, identify the changes and identify the disease of the cow. Inception V3 and the GRU (Gate Recurrent Unit) models are used to implement the video analyzing .h5 model.

Video is simply a sequence of multiple images that are updated really fast creating an appearance of a motion. Here in the study, Inception V3 approach is used to extract the features of the video frames and Gated Recurrent Network is used to capture the sequence relationship between the frames of the video.

**Table VII:** Details regarding video dataset

| Identified behavioral changes | Number of trained videos |
|---|---|
| Bovine Spongiform Encephalopathy | 90 |
| Lameness | 37 |
| Heat Stress | 21 |
| Healthy | 19 |
| Unknown | 81 |

The video dataset used in this video analyzing component is obtained through leading dairy farming countries such as India, New Zealand, Department of Agriculture, Food and Marine of Ireland by referring the websites and YouTube channels and from Sri Lankan small scale farms.

Through the surveys done, the other issue we identified is that there is no mobile application containing the facility of the prescribing the medicine by analyzing the age and the weight of the cow by referring to an image of the cow. Therefore through our study we provide ICT solution for this problem.

The age and the weight of the diseased cow is analyzed through images and according to the identified disease, the best dosage for the disease is provided. Two models were implemented to identify the cattle weight and age using Mobile Net V2 algorithm.

**Table IX:** Details regarding "age" dataset

| Identified Age Groups | Number of images trained |
|---|---|
| 1 – 5 Years | 12 |
| 5 – 10 Years | 47 |
| 11 – 15 Years | 22 |

**Table X:** Details regarding "Weight" dataset

| Identified Weight group | Number of images trained |
|---|---|
| 93 lbs. -177 lbs. | 144 |
| 183 lbs. – 278 lbs. | 80 |
| 259 lbs. – 548 lbs. | 595 |
| 498 lbs. – 738lbs. | 238 |

Dataset was collected via dairy farming sites, small scale farms in Sri Lanka and websites by going through researches.

## IV. EVALUATION AND RESULTS

### A. Breeds Identification

The breeds identification model was created using Mobile Net V2 model. 1123 image data was trained using Mobile Net V2 model.

```
Found 1123 images belonging to 5 classes.

{'Ayrshire cattle': 0,
 'Brown Swiss cattle': 1,
 'Holstein Friesian cattle': 2,
 'Jersey cattle': 3,
 'Unknown': 4}
```

**Figure 2:** Identified breeds

**Table XI:** Actual And Predicted reults of breeds

| Uploaded breed image | Actual Result | Predicted Result | Confident |
|---|---|---|---|
| Ayrshire – 1 | Ayrshire | Ayrshire | 96% |
| Ayrshire – 2 | Ayrshire | Holstein | 35% |
| Holstein | Holstein | Holstein | 99% |
| Jersey – 1 | Jersey | Brown Swiss | 20% |
| Jersey – 2 | Jersey | Jersey | 100% |
| Brown Swiss | Brown Swiss | Brown Swiss | 100% |

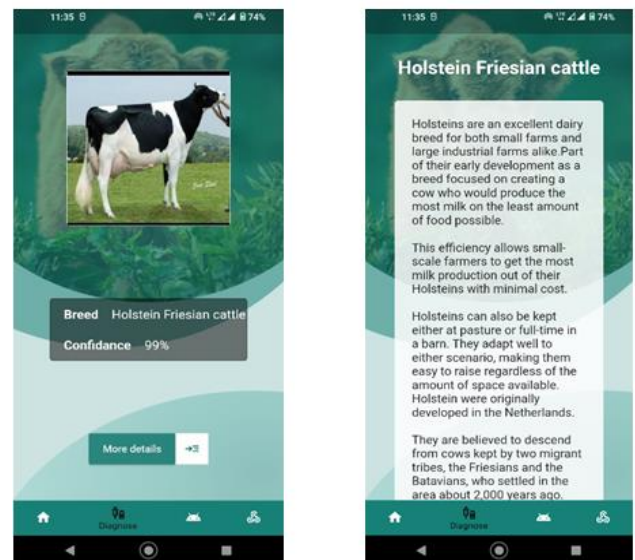

**Figure 3:** Mobile UIs of breeds identification function

### B. Diseases Identification using Images

The diseases identification model was implemented using Mobile Net V2 model by training total number of 454 image data.

```
Found 454 images belonging to 7 classes.

{'Bovine Johne_s Disease': 0,
 'Foot _ Mouth Disease': 1,
 'Healthy  Cattle': 2,
 'Lumpy Skin Disease': 3,
 'Mastitis Disease': 4,
 'Milk Fever Disease': 5,
 'Unknown': 6}
```

**Figure 4:** Trained dataset for disease identification

**Table XII:** Actual And Predicted reults of breeds

| Uploaded disease image | Actual Result | Predicted Result | Confident |
|---|---|---|---|
| Foot & Mouth | Foot & Mouth | Foot and mouth | 94% |
| Bovine John's disease | Bovine John's disease | Bovine John's disease | 99% |
| Lumpy Skin Disease | Lumpy Skin Disease | Lumpy Skin Disease | 86 % |
| Mastitis – 1 | Mastitis | Healthy | 26% |
| Mastitis – 2 | Mastitis | Mastitis | 80% |
| Milk Fever | Milk Fever | Milk Fever | 90% |
| Healthy | Healthy | Healthy | 70% |

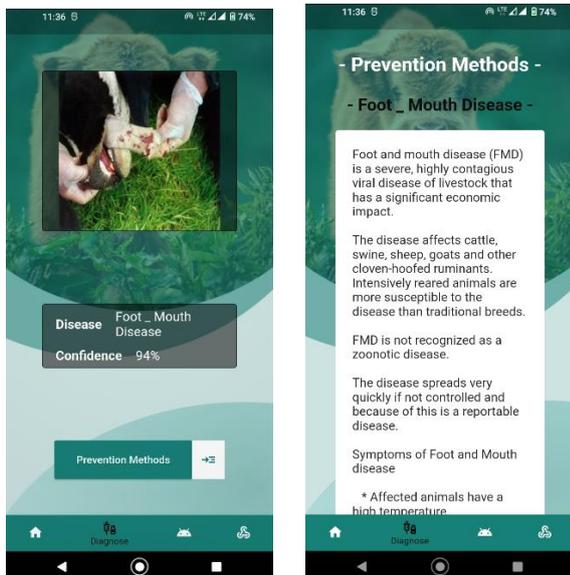

**Figure 5:** Mobile application UIs of disease identification using images function

### C. Diseases Identification using Videos

Diseases Identification using videos model was implemented using Gated Recurrent unit and Inception V3 model by training total number of 248 video files.

```
Model: "model_1"
_______________________________________________________________________________
Layer (type)              Output Shape         Param #    Connected to
===============================================================================
input_9 (InputLayer)      [(None, 200, 2048)]  0          []

input_10 (InputLayer)     [(None, 200)]        0          []

gru_2 (GRU)               (None, 200, 16)      99168      ['input_9[0][0]',
                                                           'input_10[0][0]']

gru_3 (GRU)               (None, 8)            624        ['gru_2[0][0]']

dropout_1 (Dropout)       (None, 8)            0          ['gru_3[0][0]']

dense_2 (Dense)           (None, 8)            72         ['dropout_1[0][0]']

dense_3 (Dense)           (None, 5)            45         ['dense_2[0][0]']

===============================================================================
Total params: 99,909
Trainable params: 99,909
Non-trainable params: 0
```

**Figure 6:** Video analysing model summary

**Table XIII:** Actual And Predicted reults of behavioral changes

| Uploaded video | Actual Result | Predicted result | Confident |
|---|---|---|---|
| Bovine spongiform encephalopathy | Bovine spongiform encephalopathy | Bovine spongiform encephalopathy | 90% |
| Lameness – 1 | Lameness | Lameness | 86% |
| Lameness - 2 | Lameness | Heat Stress | 45% |
| Heat Stress | Heat Stress | Heat Stress | 99% |
| Healthy | Healthy | Healthy | 70% |
| Unknown | Unknown | Unknown | 99% |

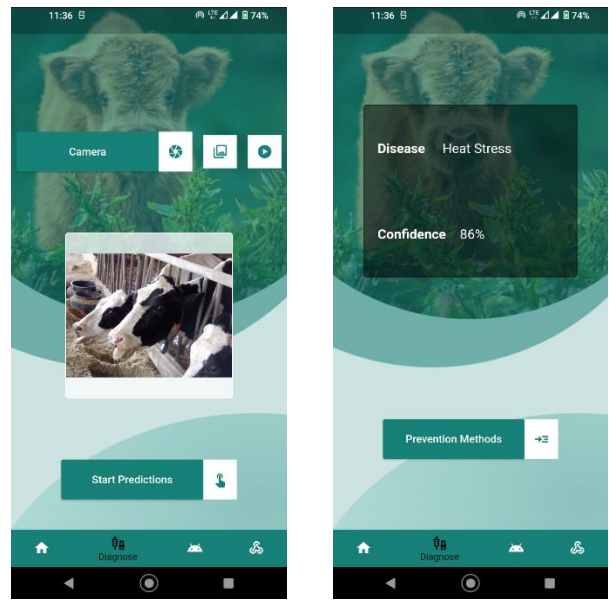

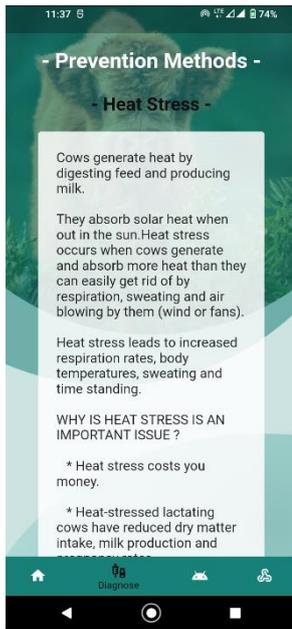

**Figure 7:** Mobile application UIs of disease identification using videos function

*D. Age and Weight Identification & Provide Customized Medicine*

      Two models were created to analyze weight and age, using Mobile Net V2 and training total number of 1356 image data.

```
Found 180 images belonging to 4 classes.
```

```
{'1 to 5 Years_Mouth': 0,
 '11to 15 Years__Mouth': 1,
 '5 to 10 Years__Mouth': 2,
 'Unknown': 3}
```

**Figure 8:** Dataset of age identification function

```
Found 1176 images belonging to 5 classes.
```

```
{'183lbs-278lbs_Body': 0,
 '259lbs-548lbs_Body': 1,
 '93lbs-177lbs_Body': 2,
 'Above 498lbs_Body': 3,
 'Unknown': 4}
```

**Figure 9:** Dataset of weight identification function

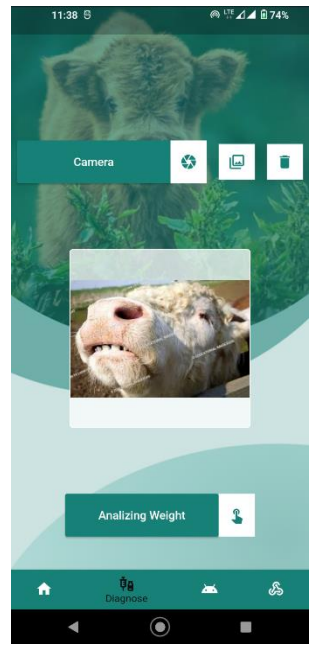
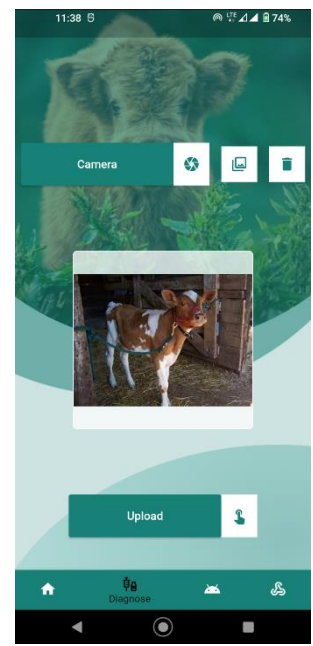
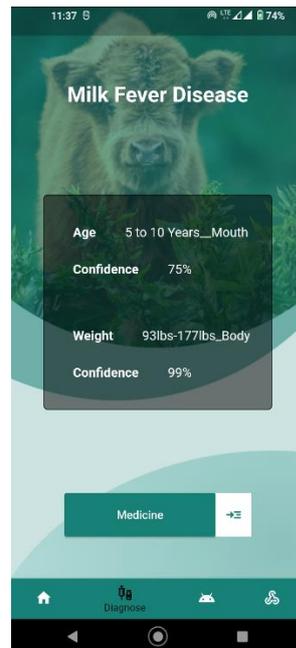
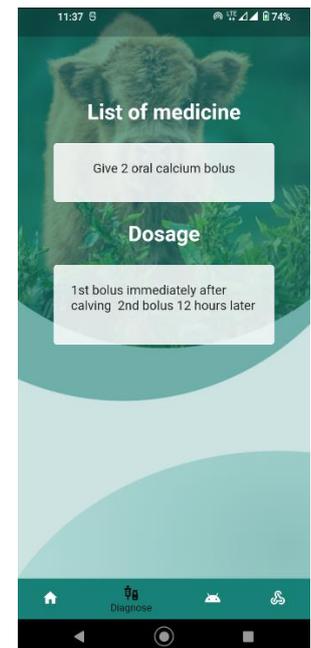

**Figure 10:** Mobile UIs of age and Weight identification Function

## V. CONCLUSION

      Our research project will save cattle population which is a great opportunity to save a valuable life. According to the above mention details , most of the cattle dies due to suffering from diseases. Our intention is to save the cattle life. Our application is easy to use by any person with any educational level .Our application is a faster method of identifying the cattle breed , identify cattle diseases and identifying the cattle age and weight. By uploading image or video identification process can be proceed. And user can experience the accuracy of the

output . All the android and iOS devices are supported which we are focusing a larger number of users for our mobile application . Therefore our application will be an useful mobile application.

## ACKNOWLEDGMENT

We are grateful for giving the guidance in each step of developing our research project from the selecting the topic ,our superior and co-superior for sharing their experience. And Abewela farm in Nuwara Eliya , veterinarians , farmers and doctors for providing necessary data for our project.